\documentclass{article}
\usepackage{bm}
\usepackage{float}
\usepackage{parskip}
\usepackage{booktabs}
\usepackage{graphicx}
\usepackage[ruled,vlined]{algorithm2e}
\usepackage[margin=3cm]{geometry}
\usepackage{amsthm}
\usepackage{ulem}
\usepackage{amsmath}
\usepackage{amsfonts}
\usepackage{amsthm}
\usepackage{caption}
\usepackage{tabularx}
\usepackage{soul}
\usepackage{authblk}
\usepackage[none]{hyphenat}
\newcolumntype{L}{>{\center\arraybackslash}X}
\usepackage[dvipsnames,svgnames,x11names]{xcolor}
\usepackage{subcaption}
\usepackage[colorinlistoftodos]{todonotes}
\usepackage{mathtools}
\usepackage[shortlabels]{enumitem}
\usepackage{fancyhdr}
\usepackage{lineno}

\newcommand{\virgolette}[1]{{``#1''}}

\usepackage[colorlinks = true,
 linkcolor = blue,
 urlcolor = blue,
 citecolor = blue,
 anchorcolor = blue]{hyperref}

\newcommand{\gra}{{\it GRAPE}}
\newcommand{\ens}{{\it Ensmallen}}
\newcommand{\emb}{{\it Embiggen}}
\newcommand{\tsne}{t-SNE}

\newcommand{\NumberOfNodeEmbeddingModels}{69}

\newcommand{\NumberOfUniqueNodeEmbeddingModels}{50}

\newcommand{\NumberOfGRAPENodeEmbeddingModels}{28}

\newcommand{\NumberOfIntegratedNodeEmbeddingModels}{41}

\DeclarePairedDelimiter\abs{\lvert}{\rvert}%

\title{
\gra~for Fast and Scalable Graph Processing and random walk-based Embedding}

\author[1]{Luca Cappelletti}
\author[1]{Tommaso Fontana}
\author[1,2,3]{Elena Casiraghi}
\author[4,5]{Vida Ravanmehr}
\author[6]{Tiffany J. Callahan}
\author[7]{Carlos Cano}
\author[3]{Marcin P. Joachimiak}
\author[3]{Christopher J. Mungall}
\author[4]{Peter N. Robinson}
\author[3]{Justin Reese}
\author[1,2,8,9,*]{Giorgio Valentini}
\affil[1]{AnacletoLab, Dipartimento di Informatica, Universit\`a degli Studi di Milano, Italy}
\affil[2]{CINI, National Laboratory in Artificial Intelligence and Intelligent Systems—AIIS, Roma, Italy}
\affil[3] {Division of Environmental Genomics and Systems Biology, Lawrence Berkeley National Laboratory, Berkeley, USA}
\affil[4]{The Jackson Laboratory for Genomic Medicine, Farmington, CT, USA}
\affil[5] {Department of Lymphoma and Myeloma, The University of Texas MD Anderson Cancer Center, Houston, TX, USA}
\affil[6]{Department of Biomedical Informatics, Columbia University Irving Medical Center, New York, USA}
\affil[7]{Department of Computer Science and Artificial Intelligence, University of Granada, Spain}
\affil[8]{European Laboratory for Learning and Intelligent Systems (ELLIS)}
\affil[9]{Data Science Research Center, Universit\`a degli Studi di Milano, Italy}
\affil[*]{Corresponding author. E-mail: valentini@di.unimi.it}

\begin{document}
\date{}
\maketitle

\clearpage
\begin{abstract}
Graph Representation Learning (GRL) methods opened new avenues for addressing complex, real-world problems represented by graphs.
However, many  graphs  used in these applications comprise millions of nodes and billions of edges and are beyond the capabilities of current methods and software implementations.
We present \gra, a software resource for graph processing and embedding that  is able to scale with big graphs by using specialized and smart data structures, algorithms, and a fast parallel implementation of random walk-based methods.
Compared with state-of-the-art software resources, \gra~shows an improvement of orders of magnitude in empirical space and time complexity, as well as competitive edge and node label prediction performance. 
\gra~comprises about 1.7 million well-documented lines of Python and Rust code and provides \(\NumberOfNodeEmbeddingModels\) node embedding methods, \(25\) inference models, a collection of efficient graph processing utilities and over \(80,000\) graphs from the literature and other sources. Standardized interfaces allow a seamless integration of third-party libraries, while ready-to-use and modular pipelines permit an easy-to-use evaluation of GRL methods, therefore also positioning \gra~as a software resource to perform a fair comparison between methods and libraries for graph processing and embedding.

\end{abstract}

\section{Introduction}

In various fields such as biology, medicine, data and network science, graphs can naturally model available knowledge as interrelated concepts, represented by a network of  nodes connected by edges. 
The wide range of graph applications has motivated the development of a rich literature on Graph Representation Learning (GRL) and inference models~\cite{hamilton2020graph}.

GRL models compute embeddings, i.e. vector representations  of the graph and its constituent elements, capturing their topological, structural, and semantic relationships.
Graph inference models can use such embeddings and available additional features for several tasks, e.g., visualization, clustering, node-label, edge-label, and edge prediction problems~\cite{hamilton2020graph}.

State-of-the-art GRL algorithms, including, among others, methods based on matrix factorization, random walks (RW), graph kernels~\cite{shervashidze2011}, triple sampling, and (deep) Graph Neural Networks (GNN)~\cite{hamilton2020graph, Wu21}, have shown  their effectiveness in analyzing networks from sociology, biology, medicine, and many other disciplines.

Although a great deal of research  has been devoted to the development of software resources for graph-processing and analysis (e.g., iGraph~\cite{Csardi06}, GraphLab~\cite{Low10}, NetworkX~\cite{Hagberg08}, GraphX~\cite{Gonzalez14} and SNAP~\cite{leskovec2016snap}), or for GRL (e.g., PecanPy~\cite{liu2021pecanpy}, PyKeen~\cite{ali2021pykeen}, DGL~\cite{wang2019deep}, PytorchGeometric~\cite{Fey/Lenssen/2019}, Spektral~\cite{grattarola2021graph}), real-world networks often include millions of nodes and billions of edges, thus raising the problem of the scalability of existing software resources~\cite{Zhang20}.
In particular, the scalability of GNNs represents an open issue~\cite{Wu21}, despite recent efforts to design GNNs that can scale with large graphs~\cite{zeng21}.

In this context, for scalability issues, RW-based GRL models are often preferred. However, their performance is often affected by the high computational costs required by the RW generators.
Indeed current state-of-the-art RW-based graph embedding libraries display a limited ability to efficiently generate enough RW data samples  to accurately represent the topology of the underlying graph.
This limits the performance of node and edge label prediction methods, which strongly depends on the informativeness of the underlying embedded graph representation. 
The efficient generation of billions of sampled RWs could lead to more accurate embedded representations of graphs and could boost the performance of machine learning methods that learn from the embedded vector representation of nodes and edges. 

The Findable, Accessible, Interoperable and Reusable (FAIR) comparison of different graph-based methods under different experimental set-ups is a relevant open issue, only very recently considered in literature in the context of the Open Graph Benchmark Large-Scale Challenge (OGB-LSC). This initiative enables a FAIR comparative evaluation of different models on three specific large-scale graphs~\cite{Hu21}. However, further efforts are required to provide standard interfaces to easily integrate methods from different libraries and public experimental pipelines, and to allow a FAIR comparison of different methods and libraries for the analysis of any graph-based data.


\gra~({\it Graph Representation leArning, Prediction and Evaluation}) provides a modular and flexible solution to the above problems by offering:  
(a) a scalable and fast software library that efficiently implements RW-based embedding methods, graph processing algorithms and inference models that can run both on general-purpose desktop and laptop computers, as well as on high-performance computing clusters; 
(b) an extensive set of efficient and effective built-in GRL algorithms that any user can continuously update by implementing easy-to-use standardized interfaces;
(c) ready-to-use evaluation pipelines to guarantee a fair and reproducible evaluation of any GRL algorithm (implemented or integrated into \gra~) using the $\sim 80,000$ graphs retrievable through the library or also other graphs provided by the user. Therefore, \gra~can also be viewed as an efficient collector of GRL methods to perform a FAIR comparison on a large set of available graphs.

%



\section{Results}

\subsection{Overview of the \gra~resource: \emb~and \ens}

\gra~ consists of about about 1.5 million lines of Python code and about 200,000 lines of Rust code  (results computed with the Tokei tool - \url{https://docs.rs/tokei/latest/tokei}), implementing efficient data structures and parallel computing techniques to enable scalable graph processing and embedding.

\begin{figure}[!tb]
 \centering
 \includegraphics[width=\textwidth]{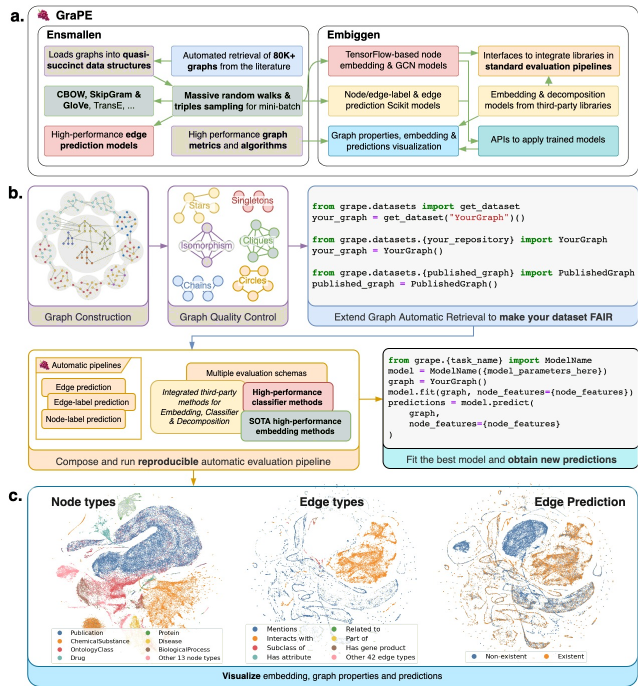}
 \caption {\textbf{Schematic diagram of \gra~(\ens~+ \emb) functionalities}. {\bf a.} High level structure of the \gra~software resource. {\bf b.} Pipelines for an easy, fair, and reproducible comparison of graph embedding techniques, graph-processing methods, and libraries. {\bf c.} Visualization of KGCOVID19 graph~\cite{Reese2021}, obtained by displaying the first two components of the \tsne~ decomposition of the embeddings computed by using a Node2Vec SkipGram model that ignores the node and edge type during the computation. The clusters' colors indicate: (left) the Biolink category for each node; (center) the Biolink category for each edge; (right) the predicted edge existence.}.
 \label{fig:schema}
\end{figure}

The library's high-level structure, overall functionalities, and its two core modules, \ens~(ENabler of SMAll computational resources for LargE Networks) and \emb~(EMBeddInG GENerator), are depicted in Fig.~\ref{fig:schema}a.

\ens~efficiently loads big graphs and executes graph processing operations, owing to its Rust~\cite{perkel2020scientists} implementation, and to the usage of   map-reduce thread-based parallelism
and branch-less Single Instruction Multiple Data (SIMD) parallelism. 
It also provides Python bindings for ease of use.

Designed to leverage succinct data structures~\cite{elias1975universal}, \gra~requires only a fraction of the memory required by other libraries and guarantees average constant-time rank and select operations~{\cite{pibiri2017dynamic}. This makes it possible to execute many graph processing tasks, e.g. accessing node neighbours and running first- and second-order RWs, with memory usage close to the theoretical minimum. 

However, the performance of RW-based embedding methods is often affected by the high computational costs required by the random-walk generators that often rely on a limited number of random-walk samples that cannot accurately represent the topology of the underlying graph. This leads to uninformative graph embeddings that affect the performance of the subsequent graph-prediction models.
To overcome these limitations \gra~ focuses on smart and efficient implementations of random-walk-based embedding methods since its main objective is to scale with large graphs (see Methods for details), while other effective but more complex models based, e.g. on GNN~\cite{Wu21} available from other libraries~\cite{Zheng20} are not yet implemented in the library, due to their well-known scaling limitations~\cite{Wu21,Happ22}.

Among the many high-performance algorithms 
implemented in \gra~, we propose an algorithm, i.e. \textit{Sorted Unique Sub-Sampling (SUSS)} that allows 
\textit{approximated} RWs to be computed to enable processing graphs that contain very high-degree nodes (degree \(> 10^6\)), unmanageable for the corresponding \textit{exact} analogous algorithms. Approximated RW can achieve edge-prediction performance comparable to those obtained by the corresponding \textit{exact} algorithm with a speed-up from two to three orders of magnitude (section~\ref{sec:methods:approximated_random_walks} and \ref{sub:Ens-rw}).


\ens~also provides many other methods and utilities, such as refined multiple holdout techniques to avoid biased performance evaluations, Bader and Kruskal algorithms for computing random and minimum spanning arborescence and connected components, stress and betweenness centrality~\cite{bader2006parallel}, node and edge filtering methods, and algebraic set operations on graphs. \ens~allows graphs to be loaded from a wide variety of node and edge list formats (section \ref{sub:fast-graph-load}). 
In addition, users can automatically load data from an ever-increasing list of over \(80,000\) graphs from the literature and elsewhere (Fig. \ref{fig:schema}b, detailed in section \ref{sub:grape-fair-data}).


\emb~provides efficient implementations of GRL and inference models~(section \ref{methods:prediction_models}), including an exhaustive set of node embedding methods, e.g., spectral and matrix factorization models such as HOPE~\cite{ou2016asymmetric}, NetMF~\cite{qiu2018network} and their variations (GLEE~\cite{torres2020glee}, SocioDim~\cite{tang2009relational} - section \ref{sec:spectral_node_embedding}).
Moreover offers from-scratch implementations of
CBOW, SkipGram and GloVe embedding methods~\cite{Mikolov13b,pennington2014glove}, that substantially outperform the Keras-based ones, as TensorFlow APIs are too coarse and high-level for such fine-grained optimizations (section~\ref{sec:random_walks_methods}). 
\gra~implements RW-based methods such as DeepWalk, Node2Vec and Walklets~\cite{grover2016node2vec,perozzi2017don}(section \ref{sec:random_walks_methods}), triple sampling methods such as LINE~\cite{tang2015line} (section \ref{sec:triples_sampling_corrupted_sampling}), and corrupted-triple sampling methods such as TransE~\cite{zhang2017} (section \ref{sec:triples_sampling_corrupted_sampling}), and, more generally, a wide range of inference methods (sections \ref{methods:prediction_models}).

\gra~provides three modular pipelines to compare and evaluate node-label, edge-label and edge prediction performance under different experimental settings (section \ref{sub:grape-fair-comparison-pipeline}, fig.~\ref{fig:schema}b), as well as utilities for graph visualization (fig.~\ref{fig:schema}c). These pipelines allow non-expert users to tailor their desired experimental setup and quickly obtain actionable and reproducible results (Fig.~\ref{fig:schema}b). Furthermore, \gra~provides interfaces to integrate third-party models and libraries (e.g., KarateClub~\cite{karateclub} and PyKeen~\cite{ali2021pykeen} libraries). This way, the evaluation pipelines can compare models implemented or integrated into \gra~ (section~\ref{sec:automated_pipeline}).

The possibility to integrate external models and the availability of graphs for testing them on the same datasets allows for answering a still open and crucial issue in literature, which regards the FAIR, objective, reproducible, and efficient comparison of graph-based methods and software implementations (Section \ref{sub:grape-fair-data}).

We used the evaluation pipelines to compare the edge  and node-label prediction performance of \(16\)  embedding models. 
Moreover, we compared \gra~ with state-of-the-art graph-processing libraries across several types of graphs having different sizes and characteristics, including big real-world graphs such as {\it Wikipedia}, the {\it CTD}, Comparative Toxicogenomic Database~\cite{Davis21} and biomedical Knowledge Graphs generated through {\it PheKnowLator}~\cite{Callahan20}, showing that \gra~achieves state-of-the-art results in processing big real-world graphs both in terms of empirical time and space complexity and prediction performance.


\subsection{Fast error-resilient graph loading}
\label{sub:fast-graph-load}

\gra~has been carefully designed to efficiently perform in space and time. In this section, we carried out a comparative study of performance with state-of-the-art graph processing libraries (including \href{https://networkx.org/}{NetworkX}~\cite{hagberg2008exploring}, \href{https://igraph.org/}{iGraph}~\cite{Csardi06}, \href{https://github.com/VHRanger/CSRGraph}{CSRGraph}, \href{https://github.com/krishnanlab/PecanPy}{PecanPy}~\cite{liu2021pecanpy}) in terms of empirical space and time used for loading \(44\) different real-world graphs (Fig.~\ref{fig:performance-comparison} a and b).
 Results show that \gra~is faster and requires less memory than state-of-the-art libraries. For instance, \gra~loads the \textit{ClueWeb09} graph (\(1.7\) billion of nodes and 8 billion of undirected edges) in less than \(10\) minutes and requires about \(60GB\) of memory, whereas the other libraries were not able to load this graph. In addition, \gra~can process many graph formats and check for common format errors simultaneously. All graphs and libraries used in these experiments are directly available from \gra. Detailed results are available in the Supplementary Information Sections 1 and 2).

\subsection{\gra~outperforms state-of-the-art libraries on RW generation} \label{sub:Ens-rw}

Through extensive use of thread and SIMD parallelism and specialized quasi-succinct data structures, \gra~outperforms state-of-the-art libraries by one to four orders of magnitude in the computation of RWs, both in terms of empirical computational time and space requirements (Figure \ref{fig:performance-comparison}-c, d, e, f and Section~\ref{sec:ensmallen_comparison}), where the method used to measure execution time and peak memory usage properly is presented in Supplementary Information Section 6.3. 

Further speed-up of second-order RW computation is obtained by dispatching one of the \(8\) optimized implementations of \textit{Node2Vec} sampling~\cite{grover2016node2vec}. The dispatching is based on the values of the \textit{return} and \textit{in-out} parameters and the type of the graph (weighted or unweighted). \gra~automatically provides the version best suited to the requested task, with minimal code redundancy (Section~\ref{methods:efficient_random_walks}). The time performance difference between the least and the most computationally expensive implementations is around two orders of magnitude (Supplementary Information Section 7.2 and Supplementary Tables 50 and 51).

\subsubsection{Experimental comparison of graph processing libraries.} \label{sec:ensmallen_comparison}

We compared \gra~with a set of state-of-the-art libraries, including \href{https://github.com/shenweichen/GraphEmbedding}{GraphEmbedding}, \href{https://github.com/eliorc/node2vec}{Node2Vec}, \href{https://github.com/VHRanger/CSRGraph}{CSRGraph} and \href{https://github.com/krishnanlab/PecanPy}{PecanPy}~\cite{liu2021pecanpy}, on a large set of first and second-order RW tasks. 
The RW procedures in the GraphEmbedding and Node2Vec libraries use the alias method (Supplementary Information Section 7.2.3). The PecanPy library also employs the alias method for small graph use-cases (less than \(10,000\) nodes). CSRGraph, on the other hand, computes the RWs lazily using Numba~\cite{lam2015numba}. Similarly, PecanPy leverages Numba lazy generation for graphs having more than \(10,000\) nodes. All libraries are further detailed in Supplementary Information Section 1.

Figure~\ref{fig:performance-comparison} shows the experimental results of a complete iteration of one-hundred step RWs on all the nodes across \(44\) graphs having a number of edges ranging from some thousands to several billion (Section~\ref{sub:fast-graph-load}).
\gra~greatly outperforms all the compared graph libraries on both first and second-order RWs in terms of space and time complexity.
Note that \gra~scales well with the biggest graphs considered in the experiments, while the other libraries either crash when exceeding 200GB of memory or take more than 4 hours to execute the task (Figure~\ref{fig:performance-comparison} c, d, e, f).

\begin{figure}
 \centering
 \includegraphics[width=0.8\textwidth]{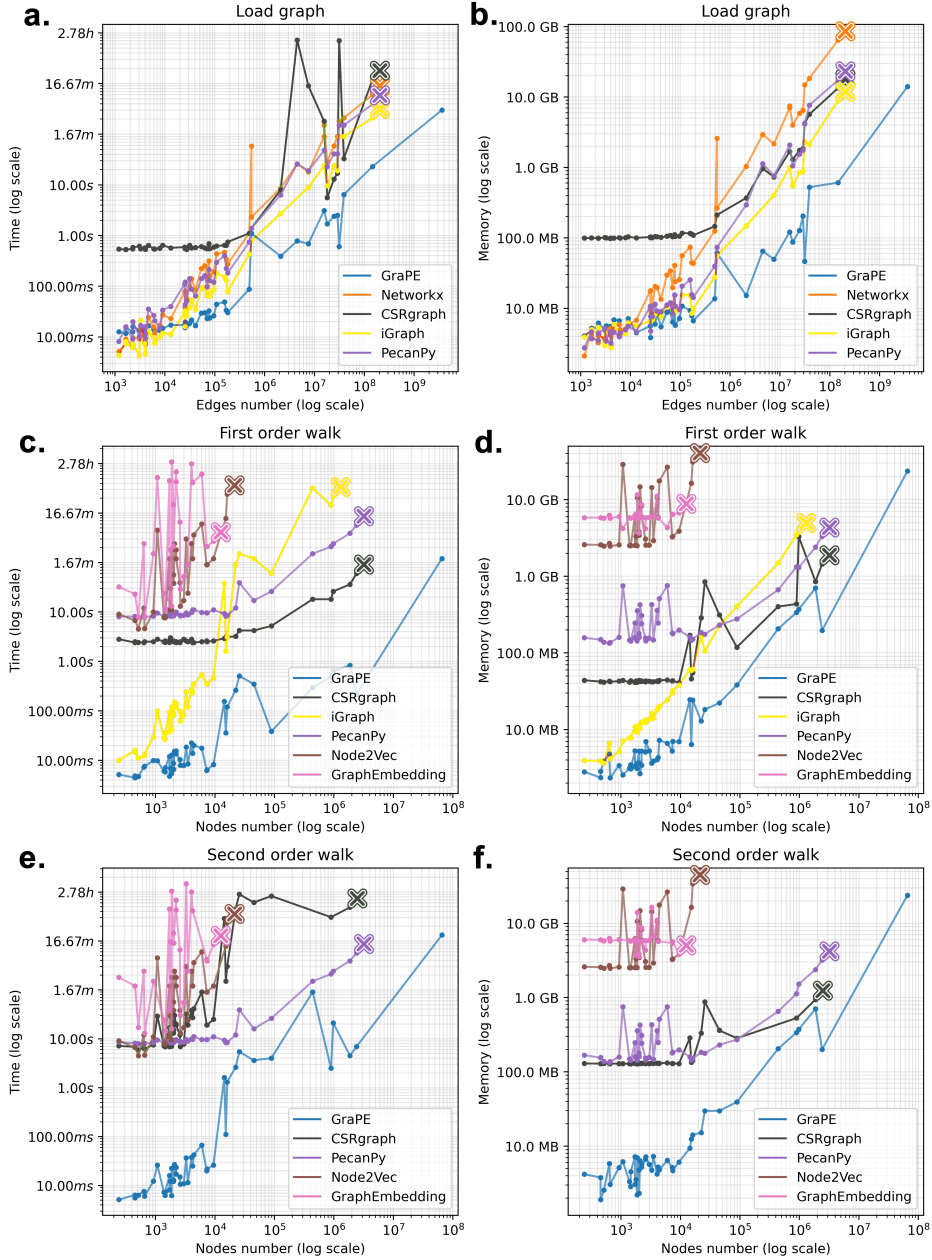}
 \caption{\small \textbf{Experimental comparison of \gra~with state-of-the-art graph processing libraries across \(44\) graphs.} 
 \textbf{Panels a and b- graph loading}: {\bf a.} Empirical execution time. {\bf b.} Peak memory usage. The horizontal axis shows the number of edges, and the vertical axis shows peak memory usage. 
 \textbf{Panels c and d - first-order RW}: {\bf c.} Empirical execution time. {\bf d.} Peak memory usage.
 \textbf{Panels e and f - second order RW}: {\bf e.} Empirical execution time. {\bf f.} Peak memory usage. The horizontal axis shows the number of nodes, and the vertical axis, respectively, execution time (c,e) and memory usage (d,f). 
 All axes are on a logarithmic scale. The \(\times\) represents when a library crashes, exceeds 200GB of memory or takes more than 4 hours to execute the task. Each line corresponds to a graph resource/library, and points on the lines refer to the \(44\) graphs used in the experimental comparison.} 
 \label{fig:performance-comparison}
\end{figure}

\subsubsection{Approximated RWs to process graphs with high-degree nodes} \label{sec:approximated_random_walks}

RWs on graphs containing high-degree nodes are challenging since multiple paths from the same node must be processed. To overcome this computational burden, \gra~provides an approximated implementation of weighted RWs that undersamples the neighbors to scale with graphs containing nodes with high-degree, e.g. with millions of neighbors (Figure~\ref{fig:approximated_random_walks} a, b, c, Section~\ref{sec:methods:approximated_random_walks}). 
To guarantee scalability, the sampling process is performed by a novel algorithm ({\it Sorted Unique Sub-Sampling - SUSS}) that we developed as an alternative to the classic and computationally demanding alias algorithm (Supplementary Information Section 7.2.3). SUSS is a sampling algorithm that divides a discrete range into \(k\) uniformly spaced buckets and randomly samples a value from each bucket to achieve an efficient neighbourhood sub-sampling for nodes with a degree \(d\gg k\). The obtained values are inherently sorted and unique (see Section~\ref{sec:methods:approximated_random_walks} for details).

We compared exact and approximated RW samples for Node2Vec-based SkipGram for edge prediction problem on the (unfiltered) {\it H. sapiens} STRING PPI network~\cite{Szklarczyk18}, achieving statistically equivalent performance (Two-sided Wilcoxon rank-sum \(p\)-value \(> 0.2\), Fig.~\ref{fig:approximated_random_walks}d), by running \(30\) holdouts, and setting a (deliberately low) degree threshold equal to \(10\) for the approximated RW, while the maximum degree in the training set ranged between $3325$ and $4184$ across the holdouts. These results show no relevant performance decay, even when using a relatively stringent degree threshold.

We used the sk-2005 graph that includes about \(50\) millions of nodes and \(1.8\) billions of edges and some nodes with degrees over \(8\) million to better show that approximated RW can be several orders of magnitude faster than the “vanilla” exact RW algorithm. Indeed by extrapolating the results reported in Fig.~\ref{fig:approximated_random_walks}e to the entire graph, the exact algorithm requires about \(23\) days, while the approximate one about \(11\) minutes, both running on a PC with two AMD EPYC 7662 64-core processors, 256 CPU threads, and 1TB RAM.

\begin{figure}[H]
 \centering
 \includegraphics[width=\textwidth]{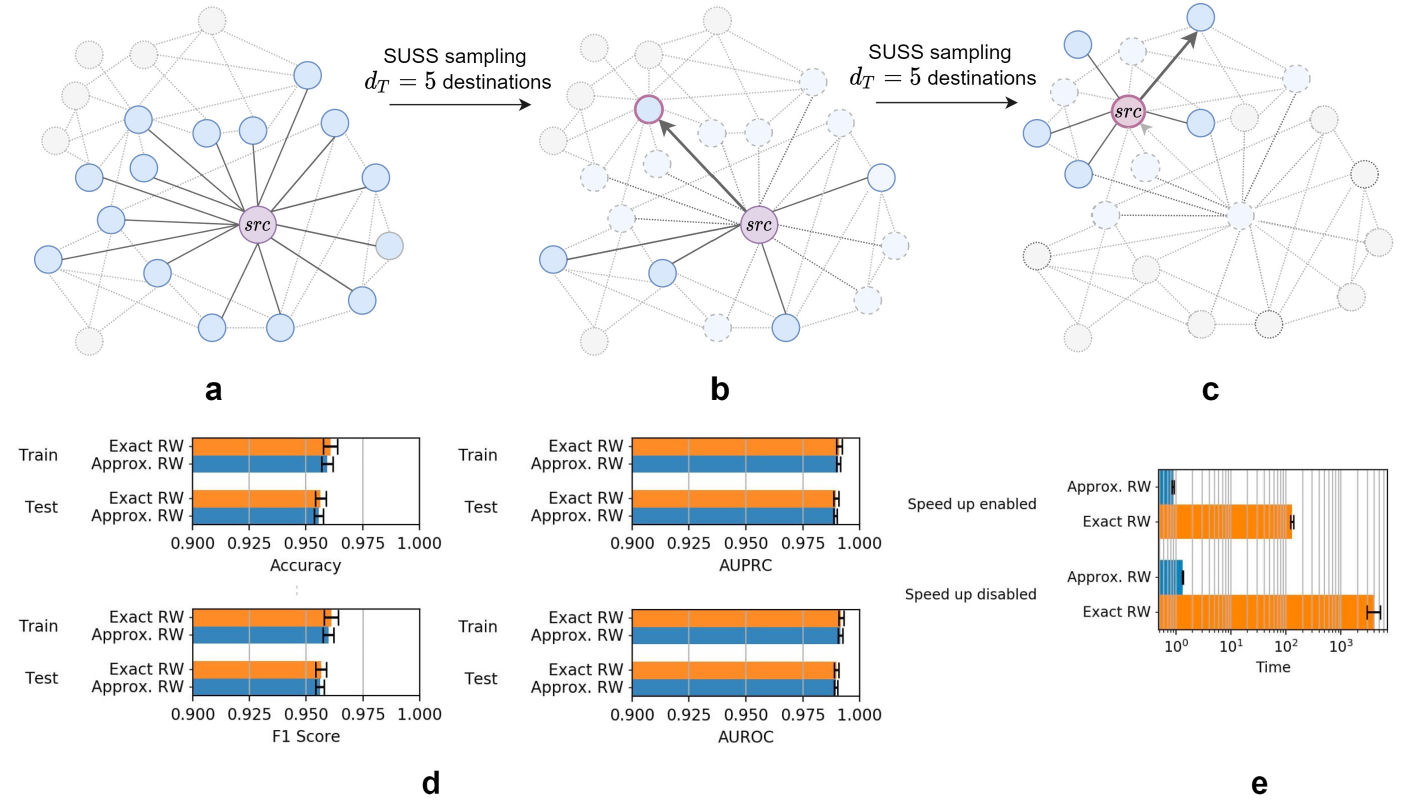}
 \caption {\textbf{Approximated RW}. \textbf{a.} The RW starts at node \(src\); its \(15\) neighbourhood nodes are highlighted in cyan. \textbf{b.} We sample \(d_T=5\) destination nodes (\(d_T\) being the degree threshold) from the available \(15\) destinations, using our Sorted Unique Sub-Sampling algorithm (SUSS, Section~\ref{sec:methods:approximated_random_walks}), and performs a random step (edge highlighted with an arrow). \textbf{c.}
 A further step will be performed on the successor node (that now becomes the novel source node \(src\)), and the same process is repeated until the end of the walk. 
 \textbf{d.} Edge prediction performance comparison (Accuracy, AUPRC, F1 Score, and AUROC computed over $n = 10$ holdouts - data are presented as mean values $+/-$ SD) using Skipgram-based embeddings and RW samples obtained with exact and approximated RWs for both the training and the test set with the STRING-PPI dataset. Bar plots are zoomed-in at \(0.9\) to \(1.0\), with error bars representing the standard deviation, computed over \(30\) holdouts.
 \textbf{e.} Empirical time comparison (in msec) of the approximated and exact second-order RW algorithm on the graph \textit{sk-2005}~\cite{BRSLLP}: \(100-steps\) RWs are run on \(100\) randomly selected nodes.  Error bars represent the standard deviation across \(n = 10\) repetitions. Time is on a logarithmic scale.
 Data are presented as mean values $+/-$ SD.}. 
 \label{fig:approximated_random_walks}
\end{figure}

\subsection{GRAPE enables a fair comparison of graph-based methods}
\label{sec:link_prediction_models}

\gra~provides both a large set of ready-to-use graphs that can be used in the experiments and standardized pipelines to fairly compare different models and graph libraries, ensuring reproducibility of the results (Fig.~\ref{fig:schema} b -- see section~\ref{sec:automated_pipeline} for details). Graph embedding is efficiently implemented in Rust from scratch (with a Python interface) or is integrated from other libraries by implementing the interface methods of an abstract \gra~class. \gra~users can compare different embedding methods and prediction models and add their own methods to the standardized pipelines. Our experiments show how to use the standardized pipelines to fairly compare a large set of methods and different implementations using only a few lines of Python code.



\paragraph{Experimental comparison of node and edge embedding methods.}
\label{sec:comparison_embedding}

We selected \(16\) among the \(\NumberOfNodeEmbeddingModels\) node embedding methods available in \gra, and we used the edge prediction and node-label standardized prediction pipelines to compare the prediction results obtained by Perceptron, Decision Tree, and Random Forest classifiers (Fig.~\ref{fig:balanced_accuracy_comparisons}). We used the Hadamard product for the edge prediction tasks to construct edge embeddings from node embeddings, i.e. the element-wise product of the source and destination nodes to obtain the embedding of the corresponding edge. We applied a \virgolette{connected Monte Carlo} evaluation schema for edge prediction and a stratified Monte Carlo evaluation schema for node-label prediction (see Supplementary Information Section 10.2 for more details).

The models have been tested on \(3\) graphs for edge prediction (Fig.~\ref{fig:balanced_accuracy_comparisons}-a,b) and \(3\) graphs for node-label prediction (Fig.~\ref{fig:balanced_accuracy_comparisons}-c,d). The graph reports, describing the characteristics of the analyzed graphs, automatically generated with \gra, are available in the Supplementary Information Sections 3.2 and 3.3. Since they are homogeneous graphs (i.e. graphs having only one type on nodes and edges) we considered only homogeneous node embedding methods. Moreover, we discarded non-scalable models, e.g. models based on the factorization of dense adjacency matrices. 

Among the \(16\) methods, \(11\) are implemented in \gra~(purple in Fig.~\ref{fig:balanced_accuracy_comparisons}) and \(5\) have been integrated from the Karate Club library~\cite{karateclub} (cyan in Fig.~\ref{fig:balanced_accuracy_comparisons}). They can be grouped into three broad classes:
\begin{description}
\item[a. Spectral and matrix factorization methods:] Geometric Laplacian Eigenmap Embedding (GLEE)~\cite{torres2020glee}, Alternating Direction Method of Multipliers for Non-Negative Matrix Factorization (NMFADMM)~\cite{sun2014alternating}, High-Order Proximity preserved Embedding (HOPE)~\cite{ou2016asymmetric}, Iterative Random Projection Network Embedding (RandNE)~\cite{zhang2018billion}, Network Matrix Factorization (NetMF)~\cite{qiu2018network}, and Graph Representations (GraRep)~\cite{cao2015grarep}.
\item[b. First-order random-walk methods:] DeepWalk-based GloVe, CBOW, and SkipGram, Walklets SkipGram~\cite{perozzi2017don,pennington2014glove,Mikolov13b}, and Role2Vec with Weisfeiler-Lehman Hashing~\cite{ahmed2019role2vec,shervashidze2011,karateclub}.
\item[c. Second-order RW methods:] Node2Vec-based GloVe, CBOW, and SkipGram~\cite{pennington2014glove,Mikolov13b,grover2016node2vec}.
\item[d. Triple-sampling methods:] first and second order LINE~\cite{tang2015line}.

\end{description}

All the embedding methods and classifiers are described in more detail in sections \ref{sec:spectral_node_embedding}, \ref{sec:random_walks_methods}, \ref{sec:triples_sampling_corrupted_sampling}, and \ref{methods:prediction_models}.

Results show that no model is consistently better with respect to the others across the types of tasks and the data sets used in the experiments (Figure~\ref{fig:balanced_accuracy_comparisons}). These results are analogous to those obtained by Kadlec et al.~\cite{kadlec-etal-2017-knowledge} for the TransE model family and those obtained by Errica et al. ~\cite{errica_fair_2020} for GNN models, highlighting the need for objective pipelines to systematically compare a wide array of possible methods for a desired task. 
The standardized pipelines implementing the experiments are available from the online \gra~tutorials and allow the full reproducibility of the results summarized in Fig.~\ref{fig:balanced_accuracy_comparisons}. Full results using other evaluation metrics are available in Supplementary Information Sections 5.1 and 5.2.

\begin{figure}
    \centering
    \includegraphics[width=0.8\textwidth]{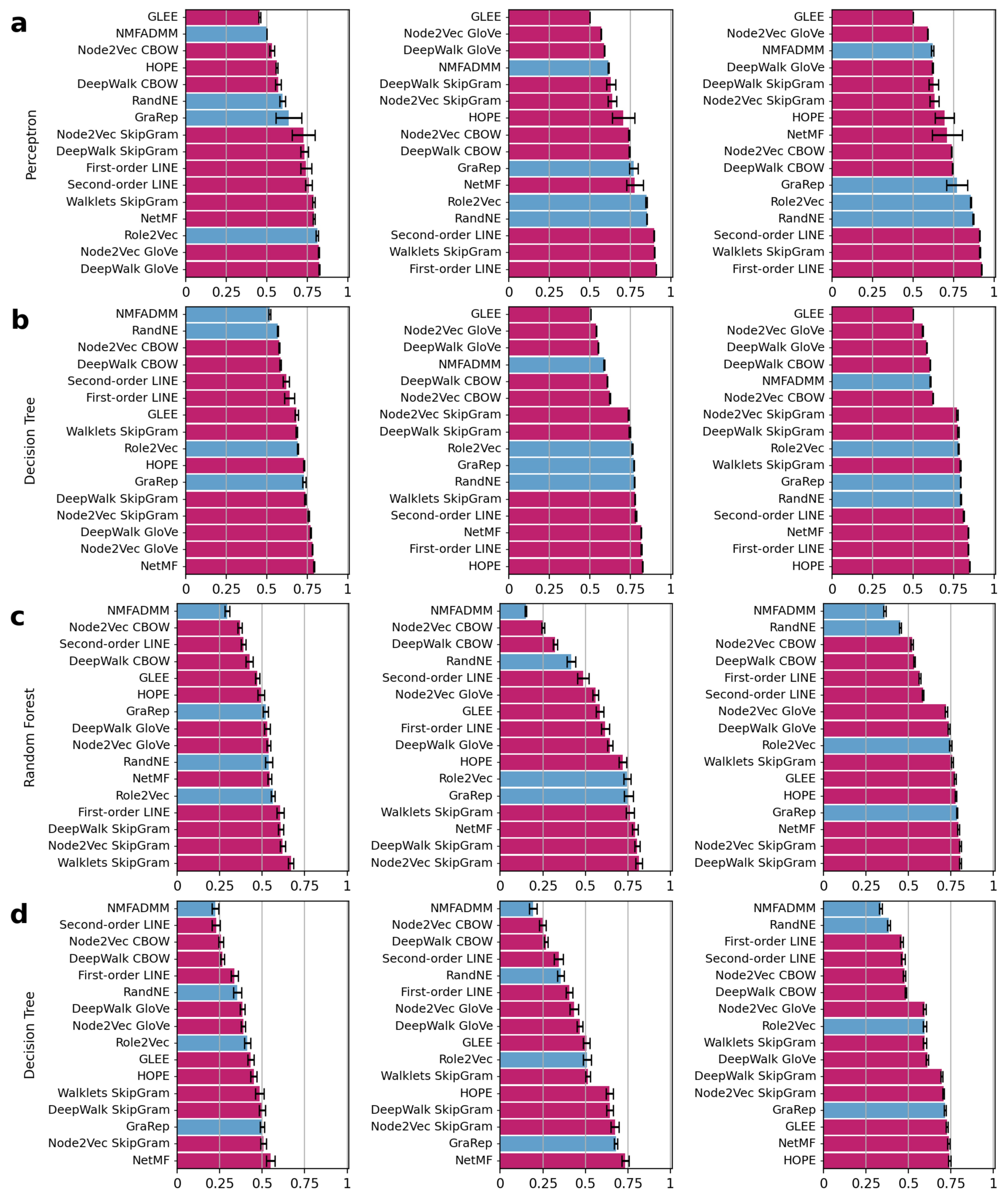}
    \caption{\small \textbf{Comparison of embedding methods through the \gra~pipelines: edge and node label prediction results.}
    Results represent the mean balanced accuracy computed across $n = 10$ holdouts $+/-$ SD (results using other evaluation metrics are available in Supplementary Information Section 5). We sorted the embedding models by performance for each task; methods directly implemented in \gra~are in purple, while integrated methods are in cyan.
    \textbf{(a, b)}: Edge prediction results obtained through a Perceptron (a) and a Decision tree (b). Barplots from left to right show the balanced accuracy results obtained with the \textit{Human Phenotype Ontology} (left), STRING \textit{Homo sapiens} (center) and STRING \textit{Mus musculus} (right).
    \textbf{(c, d)}: Node-label prediction results obtained through a Random Forest (c) and a Decision Tree (d). Barplots from left to right show the balanced accuracy respectively achieved with \textit{CiteSeer} (left), \textit{Cora} (center) and \textit{PubMed Diabetes} (right) datasets.}
  \label{fig:balanced_accuracy_comparisons}
\end{figure}

\subsection{Scaling with big real-world graphs} \label{sub:grape-scaling}

To show that \gra~can scale and boost edge prediction in big real-world graphs, we compared its node2vec-based models with state-of-the-art implementations on three big graphs: 1) English Wikipedia graph; 2) a graph constructed using the Comparative Toxicogenomic Database (CTD~\cite{Davis21}); 3) A biomedical graph generated through PheKnowLator~\cite{Callahan20}.
Supplementary Information Section 6.1 reports details about the construction and the characteristics of the three graphs.

\subsubsection{Experimental set-up} 
\label{subsub:exp-setup}
In the experiments, the \gra~implementation of node2vec with both CBOW and SkipGram were compared with those available in the following embedding libraries, widely used by the scientific community: \textit{PecanPy} \cite{liu2021pecanpy}, \textit{NodeVectors} (\url{https://github.com/VHRanger/nodevectors}), \textit{SNAP} \cite{leskovec2016snap}, \textit{Node2Vec} (\url{https://github.com/eliorc/node2vec}), \textit{GraphEmbedding} (\url{https://github.com/shenweichen/GraphEmbedding}), \textit{FastNode2Vec} (\url{https://github.com/louisabraham/fastnode2vec}) and \textit{PyTorch Geometric} (\url{https://github.com/pyg-team/pytorch_geometric}). 
More details about the above state-of-the-art libraries are reported in Supplementary Information Section 6.2.

The embeddings computed by each of the tested models were used to train 
a Decision tree available from the {\it Embiggen}~module of \gra~ for edge prediction. To perform an unbiased evaluation, training and tests were performed by $10$ connected Monte Carlo holdouts (with a $80:20$ train:test ratio - Supplementary Information Section 10.2) and performances were evaluated by precision, recall, accuracy, balanced accuracy, F1, AUROC, and AUPRC. In the experimental set-up we imposed the following memory and time constraints, using a Google Cloud VM with 64 cores and N1 Cpus with Intel Haswell micro-architecture: 
\begin{itemize}
 \item A maximum time of 48 hours for each holdout to produce the embedding;
 \item A 64GB maximum memory usage allowed during the embedding.
 \item A 256GB maximum memory usage allowed during the prediction phase.
\end{itemize}

\subsubsection{Results on scaling tests} \label{subsub:node2vec_comparisons}

\begin{figure}
 \centering
 \includegraphics[width=\textwidth]{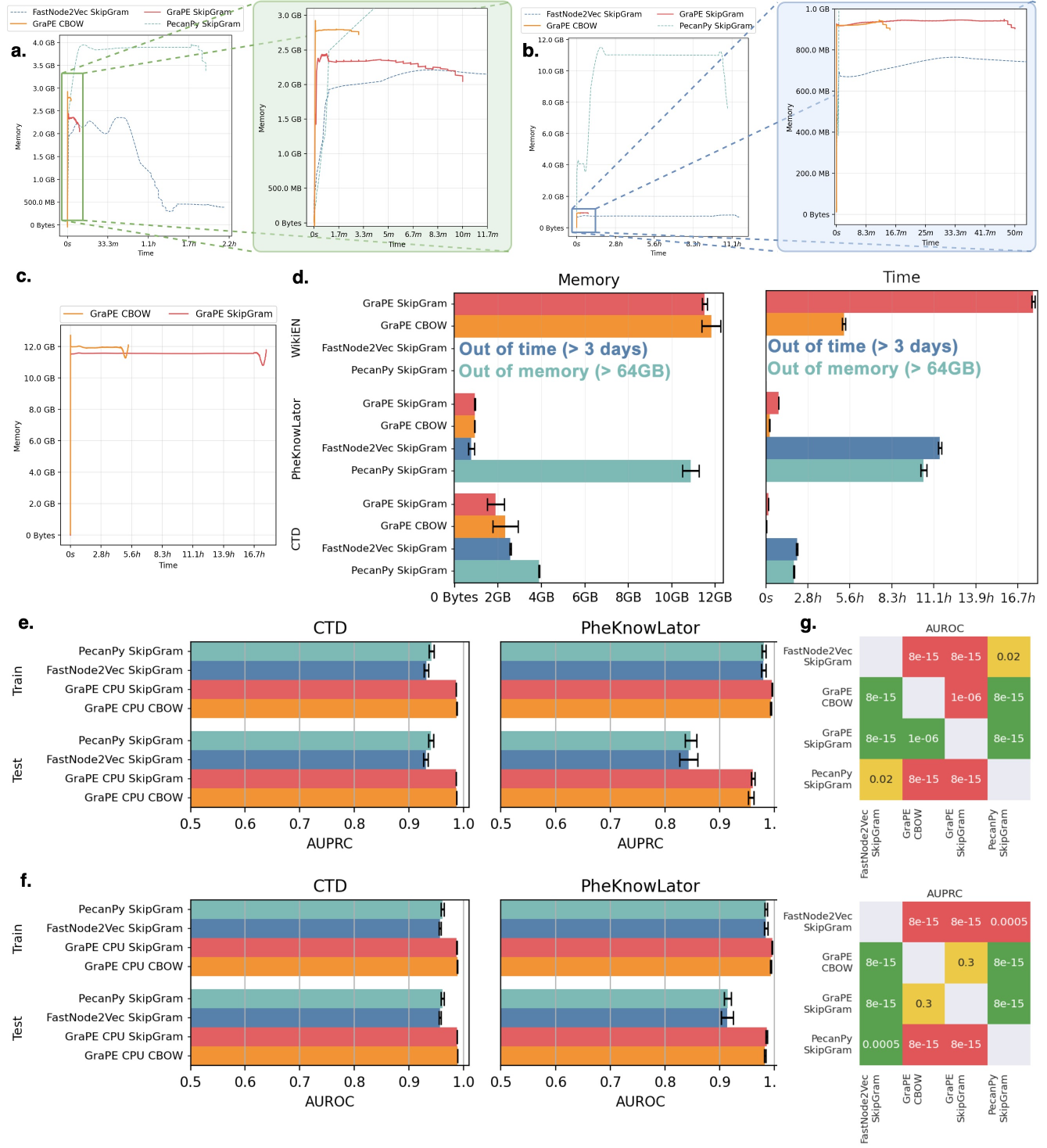}
 \caption{\footnotesize \textbf{Performance comparison  between \gra~and state-of-the-art implementations of Node2Vec on real-world big graphs.} \gra~implementations achieve substantially better empirical time complexity: (\textbf{a.}), (\textbf{{b.}}) and (\textbf{c.})  show  the worst performance (maximum time and memory, denoised using a Savitzky–Golay filter) over \(10\) holdouts on \textit{CTD}, \textit{PheKnowLator} and \textit{Wikipedia}, respectively. In \textbf{a.} and \textbf{b.} the rectangles in the left figure are magnified in the right figure to highlight \gra~ performances.
 In the \textit{Wikipedia} plot (\textbf{c.}) only \gra~results are available as the others either go out-of-time or out-of-memory. (\textbf{d.}) Average memory and computational time across $n = 10$ holdouts; data are presented as mean values $+/-$ SD.
 (\textbf{e.}) AUPRC and (\textbf{f.}) AUROC results of Decision Trees trained with different graph embedding libraries - data are presented as mean values +/- SD computed over $n= 10$ holdouts: \gra~embedding achieve better edge prediction performance than those obtained by the other libraries. (\textbf{g.}) One-sided Wilcoxon signed-rank tests results (p-values) between \gra~ and the other state-of-the-art libraries, where the win of a row against a column is in green, the tie in yellow, and the loss in red).} 
 \label{fig:benchmarks_large_graphs}
\end{figure}

\paragraph{\textit{GRAPE} can scale with big graphs when the other competing libraries fail.}

Most competing libraries could not complete the embedding and prediction tasks on big real-world graphs. Indeed \textit{NodeVectors} exceeded the time computation limit, while \textit{SNAP}, \textit{Node2Vec}, \textit{GraphEmbedding}, and \textit{PyTorch Geometric} went out of memory in the embedding phase, exceeding the available RAM memory (\(64GB\)), while \gra~only required \(54MB\) with the CTD graph. 
For the first three libraries, this is due to the extremely high memory complexity required by the Alias method they use for pre-computing the transition probabilities (Supplementary Information  Section 7.2.3); indeed the Alias method has quadratic complexity with respect to the number of nodes in the graph, therefore becoming quickly too expensive on big graphs. 
We also ran \textit{PyTorch Geometric} on a substantially smaller graph (the STRING {\it Homo sapiens} graph, having about $20$K nodes and $12M$ edges) and we registered that \gra~is about $60$ time faster than \textit{PyTorch Geometric}.

Such comparison is impossible with the other three libraries employing the Alias method, as this smaller graph is still significantly larger than what is possible to handle with them.
\textit{FastNode2Vec} and \textit{PecanPy} went out of time (more than \(48\)h of computation) on the biggest Wikipedia graph. In practice, only \gra~was able to successfully terminate the embedding and prediction tasks with all three big real-world graphs.

\paragraph{\textit{GRAPE} improves upon the empirical time complexity of state-of-the-art libraries.}

Fig.~\ref{fig:benchmarks_large_graphs} a, b and c show the memory and time requirements of \gra, \textit{FastNode2Vec} and \textit{PecanPy} (note that the other state-of-the-art libraries \ ran out of time or  memory on these real-world graph prediction tasks. With \textit{CTD} and \textit{PheKnowLator} biomedical graphs we can observe a speed-up of about one order of magnitude (Fig.~\ref{fig:benchmarks_large_graphs} a, b) of \gra~with respect to both \textit{FastNode2Vec} and \textit{PecanPy} with also a significant gain in memory usage with respect to \textit{PecanPy} and a comparable memory footprint with \textit{FastNode2Vec}. These results are confirmed by the average memory and time requirements across ten holdouts (Fig.~\ref{fig:benchmarks_large_graphs} d). Note that both \textit{FastNode2Vec} and \textit{PecanPy} fail with the Wikipedia task, while \gra~was able to terminate the computation in a few hours using a reasonable amount of memory (Fig.~\ref{fig:benchmarks_large_graphs} c and d).

\paragraph{\textit{GRAPE} boosts edge prediction performance.}
\gra~not only allows graph embedding approaches to be applied to graphs that are bigger than what was previously possible and enables fast and efficient computation, but can boost prediction performance on big real world graphs. 
Fig.~\ref{fig:benchmarks_large_graphs}-e and f show that \gra~achieves better results on edge prediction tasks with both \textit{CTD} and \textit{PheKnowLator} biomedical graphs. \gra~outperforms the other competing libraries at \(0.01\) significance level, according to the Wilcoxon rank sum test (Fig.~\ref{fig:benchmarks_large_graphs} g).
The edge embeddings have been used to train a decision tree to allow a safe comparison between the embedding libraries.

Supplementary Information  Section 6.4 reports AUROC, accuracy, and F1-score performances and other more detailed results about the experimental comparison of \gra~with state of the art libraries.

\section{Discussion}

We have presented \gra, a software resource with specialized data structures, algorithms, and fast parallel implementations of graph processing methods coupled with efficient implementations of algorithms for RW-based GRL. Our experiments have shown that \gra~significantly outperforms state of the art graph processing libraries in terms of empirical space and time complexity, with an improvement of up to several orders of magnitude for common RW-based analysis tasks. This allows substantially bigger graphs to be analyzed and may improve the performance of graph ML methods by allowing for more comprehensive training, as shown by our experiments performed on three real-world large graphs.
In addition, the substantial reduction of the computational time achieved by \gra~in common graph processing and learning tasks, will help  to reduce the carbon footprint of ML researchers and graph processing and analyzing practitioners in several disciplines.

Thanks to (1) the huge number of well-known graphs that can be efficiently loaded and used via \gra, (2) the standard interfaces that allow any user to integrate their own GRL models into \gra, and (3) the modular pipeline that allows to easily design different benchmarking experiments,  \gra~can be considered as the first resource that truly allows a FAIR comparison between virtually any method and using any graph data (including graph data directly provided by the users).

Indeed, to our knowledge the only related resource that allows a related comparison is the OGB resource~\cite{Hu21}; however 
as witnessed by the recent OGB-LSC~(\url{https://ogb.stanford.edu/neurips2022/}), the datasets and the organization of the OGB resource are well-suited for specific large-scale challenges, while the \gra~evaluation pipelines are useful to assess and compare any method on any graph benchmark chosen by any user. This makes the two resources related but complementary in their different purposes.

We would further remark that \gra~currently provides efficient implementations of RW-based embeddings, whose advantage is their applicability to a larger set of learning problems since the computed embeddings are usually task-independent and unsupervised.
Opposite to them, embeddings computed by GNNs are task-dependent and supervised, and their application to graphs with thousands of nodes and millions of edges is still hampered by GNN scalability issues that represent an open research question in literature.
For this reason, future works will be aimed at investigating how to efficiently implement deep GNN to obtain deep neural models able to efficiently scale with very big graphs~\cite{zeng21,Happ22}. More precisely, even if Elias-Fano based data structures and the SUSS algorithm proposed in this paper have been designed to efficiently implement RW embedding methods, in future research we plan to consider their integration in the context of GNNs. 

Considering the ever-increasing amount of knowledge graphs being constructed in several disciplines, \gra~may be considered as a powerful, effective, and efficient resource for advancing knowledge by performing graph-inference tasks to uncover hidden relationships between concepts, or to predict properties and discover structures in complex graphs. However, a limitation of the current implementation of \gra~is the limited availability of algorithms specifically designed for the analysis of heterogeneous graphs, but we are already working to fill this gap.

\gra~focuses primarily on CPU models, since most existing GPU VRAM are too small for several real-world graphs, leading to latency problems as data are moved back and forth between RAM and VRAM. Recently introduced top-tier GPU models provide VRAM which is considerably larger than previously available ones, potentially making it viable to translate the current CPU implementation into a GPU implementation.

Though \gra~allowed to compare different experimental setups by composing experiments on different graphs, and by using several embedding methods and prediction models, no method systematically outranked other models (Section~\ref{sec:comparison_embedding}). To close this knowledge gap, in future work we plan to run \gra~with a large scale grid-search to identify task-specific trends for the various combinations of models and their parameters.

\clearpage
\section{Methods}\label{sec:methods}

\gra~provides a wide spectrum of graph processing methods, implemented within the \ens~module, including node embedding methods (sections \ref{sec:spectral_node_embedding}, \ref{sec:random_walks_methods}, \ref{sec:triples_sampling_corrupted_sampling}), methods to combine the node embeddings for obtaining edge embeddings (section \ref{methods:edge_embedding_models}), and models for node-label, edge-label and edge prediction (section \ref{methods:prediction_models}), implemented within the \emb~module. The graph processing methods include fast graph loading, multiple graph holdouts, efficient first and second-order RWs, triple and corrupted-triple sampling, plus a wide range of graph processing algorithms that nicely scale with big graphs, using parallel computation and efficient data structures to speed up the computation.

\ens~is implemented using Rust, with fully documented Python bindings. Rust is a compiled language gaining importance in the scientific community~\cite{perkel2020scientists} thanks to its robustness, reliability, and speed. Rust allows threads and data parallelism  to be exploited robustly and safely. To further improve efficiency, some core functionalities of the library, such as the generation of pseudo-random numbers and sampling procedures from a discrete distribution, are written in assembly (see Supplementary Information Sections 7.2.1 and 7.2.2).

\gra~currently provides \(\NumberOfUniqueNodeEmbeddingModels\) unique node embedding models (\(\NumberOfNodeEmbeddingModels\) considering redundant implementations, important for benchmarks), with \(\NumberOfGRAPENodeEmbeddingModels\) being \virgolette{by-scratch} implementations and \(\NumberOfIntegratedNodeEmbeddingModels\) integrated from third-party libraries. The list of available node embedding methods is constantly growing, with the ultimate goal to provide a complete set of efficient node embedding models. The input for the various models (e.g. RWs and triples) are provided by \ens~in a scalable, highly efficient, and parallel way (Fig.~\ref{fig:schema}a). All models were designed according to the ``composition over inheritance'' paradigm, to ensure a better user experience through increased modularity and polymorphic behaviour~\cite{gamma1995elements}. More specifically, \emb~provides interfaces, specific for either the embedding or each of the prediction-tasks, that must be implemented by all models; third-party models, such as PyKeen~\cite{ali2021pykeen}, KarateClub~\cite{karateclub} and Scikit-Learn~\cite{scikit-learn} libraries, are already integrated within \gra~by implementing these interfaces. \gra~users can straightforwardly create their models and wrap them by implementing the appropriate interface.

\gra~has a comprehensive test suite. However, to thoroughly test it against many scenarios, we also employed fuzzers, that is tools that iteratively generate inputs to find corner cases in the library.

In the next section we describe the succinct data structures used in the library and detail their efficient \gra~implementation (Section~\ref{sec:data-structure}). We then summarize the spectral and matrix factorization (Section~\ref{sec:spectral_node_embedding}), the RW-based (Section~\ref{sec:random_walks_methods}), the triple and corrupted triples-based (Section \ref{sec:triples_sampling_corrupted_sampling}) embedding methods and their ~\gra~ implementation. In section \ref{methods:edge_embedding_models} we describe the edge embedding methods and in Section \ref{methods:prediction_models} the node and edge label prediction methods available in \gra. Finally in Section \ref{sec:automated_pipeline} we detail the \gra~standardized pipelines to evaluate and compare models for graph prediction tasks.

\subsection{Succinct data structures for adjacency matrices}
\label{sec:data-structure}

Besides heavy exploitation of parallelism, the second pillar of our efficient implementation is the careful design of the data structures for using as little memory as possible and quickly performing operations on them. The naive representation of graphs explicitly stores its adjacency matrix, with a \(\mathcal{O}(|V|^2)\) time and memory complexity, being \(|V|\) the number of nodes, which leads to intractable memory costs on large graphs. However, since most large graphs are highly sparse, this problem can be mitigated by storing only the existing edges. 
Often, the adopted data structure is a Compressed Sparse Rows matrix (CSR~\cite{saad2001parallel}), which stores the source and destination indices of existing edges into two sorted vectors. In \ens~we further compressed the graph adjacency matrix by adopting the Elias-Fano succinct data scheme \cite{elias1975universal}, to efficiently store the edges (Supplementary Information Section 7.1). Since Elias-Fano representation stores a sorted set of integers using memory close to the information-theoretical limit, we defined a bijective map from the graph-edge set and a sorted integer set. To define such encoding, we assigned a numerical id from a dense set to each node, and then we defined the encoding of an edge as the concatenation of the binary representations of the numerical ids of the source and destination nodes. This edge encoding has the appealing property of representing the neighbours of a node as a sequential and sorted set of numeric values, and can therefore be employed in the Elias-Fano data structure. Elias-Fano has faster sequential access than random access (Supplementary Information Section 7.1.1) and is well suited for graph processing tasks such as retrieving neighbours during RW computation and executing negative sampling using the outbound or inbound node degrees scale-free distributions.
\gra~provides both CSR and Elias-Fano based data structures for graph representation to allow a  time/memory complexity trade-off for processing large graphs.

\subsubsection{Memory Complexity}
Elias-Fano is a quasi-succinct data representation scheme, which provides a memory efficient storage of a monotone list of \(n\) sorted integers, bounded by \(u\), by using at most \(\mathcal{EF}(n, u) = 2n + n \left \lceil \log_2 \frac{u}{n} \right \rceil\) bits, which was proven to be less than half a bit per element away from optimality~\cite{elias1975universal} and assures random access to data in average constant-time.
Thus, when Elias-Fano is paired with the previously presented encoding, the final memory complexity to represent a graph \(G(V, E)\) is \(\mathcal{EF}_\phi(|V|, |E|) = \mathcal{O} \left( |E| \log \frac{|V|^2}{|E|} \right)\); this is asymptotically better than the \(\mathcal{O}\left( |E| \log |V|^2 \right)\) complexity of the CSR scheme.

\subsubsection{Edge Encoding} 
\ens~converts all the edges of a graph \(G(V, E)\) into a sorted list of integers. Considering an edge $e = (v, x) \in E$ connecting nodes $v$ and $x$ represented with, respectively, integers $a$ and $b$, the binary representation of $a$ and $b$ are concatenated through the function \(\phi_k(a, b)\) to generate an integer index uniquely representing the edge $e$ itself:
\[\phi_k(a, b) = a~2^{k} + b, \text{ where } k = \lceil \log_2 |V| \rceil \qquad\Rightarrow\qquad a = \left \lfloor \frac{\phi_k(a, b) - b}{2^k} \right \rfloor, \qquad b = \phi_k(a, b) - a~2^k\]
This implementation is particularly fast because it requires only few bit-wise instructions:
\[\phi_k(a, b) = a << k | b \qquad\Rightarrow\qquad a = \phi_k(a, b) >> k ,\qquad b = \phi_k(a, b)~\&~(2^k- 1)\]
where \(<<\) is the left bit-shift, \(|\) is the bit-wise OR and \(\&\) is the bit-wise AND (see Supplementary Information Section 7.1.1 for an example and an implementation of the encoding). Since the encoding uses \(2k\) bits, it has the best performances when it fits into a CPU word, which is usually 64-bits on modern computers, meaning that the graph must have less than \(2^{32}\) nodes and and less than \(2^{64}\) edges. However, by using multi-word integers it can be easily extended to even larger graphs.

\subsubsection{Operations on Elias-Fano.} 
The aforementioned encoding, when paired with Elias-Fano representation, allows an even more efficient computation of random-walk samples. Indeed, Elias-Fano representation allows performing \textbf{rank} and \textbf{select} operations by requiring on average constant time. These two operations were initially introduced by Jacobson to simulate operations on general trees, and were subsequently proven fundamental to support operations on data structures encoded through efficient schemes. In particular, given a set of integers \(S\), Jacobson defined the \textbf{rank} and \textbf{select} operations as follows~\cite{pibiri2017dynamic}:
\begin{flalign*}
 \mathbf{rank}(S,m) &\text{\qquad returns the number of elements in \(S\) less or equal than \(m\)}\\
	\mathbf{select}(S,i) &\text{\qquad returns the \(i\)-th smallest value in \(S\)}
\end{flalign*}
As explained below, to speed up computation, we deviate from this definition by defining the rank operation as the number of elements strictly lower than \(m\).
To compute the neighbours of a node using the rank and select operations, we observe that for every pair of nodes \(\alpha, \beta\) with numerical ids \(a, b\) respectively, it holds that:
\[a~2^{k} \le a~2^{k} + b < (a+1)~2^{k} \qquad\Rightarrow\qquad \phi_k(a, 0) \le \phi_k(a, b) < \phi_k(a + 1, 0) \]
Thus, the encoding of all the edges with source \(\alpha\) will fall in the discrete range 
\[\Big [\phi_k(a, 0), ~\phi_k(a+1, 0) \Big) = \Big [ a~2^{k}, ~(a+1)~2^{k} \Big )\]
Thanks to our definition of the \textbf{rank} operation and the aforementioned property of the encoding, we can easily derive the computation of the degree \(d(a)\) of any node $v$ with numerical id \(a\) for the set of encoded edges \(\Gamma\) of a given graph, which is equivalent to the number of outgoing edges from that node:
\[d(a) = \mathbf{rank}(\Gamma, \phi_k(a + 1, 0)) - \mathbf{rank}(\Gamma, \phi_k(a, 0))\]
Moreover, we can retrieve the encoding of all the edges $\Gamma_{a}$ starting from $v$ encoded as $a$, by selecting every index value \(i\) falling in in the range 
$[\phi_k(a, 0),\phi_k(a + 1, 0)$:
\[\Gamma_{a} = \Big \{\mathbf{select}(\Gamma, i) \mid \mathbf{rank}(\Gamma, \phi_k(a, 0)) \le i < \mathbf{rank}(\Gamma, \phi_k(a + 1, 0)) \Big \} \]
We can then decode the numerical id of the destination nodes from $\Gamma_{a}$, thus finally obtaining the set of numerical ids of the neighbours nodes \(N(a)\):
\[N(a) = \Big \{ \mathbf{select}(\Gamma, i) \;\& (2^k- 1) \mid \mathbf{rank}(\Gamma, \phi_k(a, 0)) \le i < \mathbf{rank}(\Gamma, \phi_k(a + 1, 0)) \Big \}\]

In this way, by exploiting the above integer encoding of the graph and the Elias-Fano data scheme, we can efficiently compute the degree and neighbours of a node using rank and select operations.

\subsubsection{Efficient implementation of Elias-Fano.}
The performance and complexity of Elias-Fano heavily relies on the details of its implementation. In this section our implementation is sketched, to show how we obtain an average constant time complexity for rank and select operations. A more detailed explanation can be found in the Supplementary Information Section 7.1. 

Elias-Fano is essentially aimed at the efficient representation of a sorted list of integers \(y_0, ..., y_n \) bounded by \(u\), i.e. \( \forall i \in \{1, \ldots, n-1\}\) it represents \(0 \le y_{i-1} \le y_i \le y_{i+1} \le u\). 

To this aim, it initially splits each value, \(y_i\), into a low-bits, \(l_i\), and a high-bits part, \(h_i\), where it can be proven that the optimal split between the high and low bits requires \(\left \lfloor \log_2 \frac{u}{n} \right \rfloor\) bits~\cite{pibiri2017dynamic}. 

The lower-bits are consecutively stored into a low-bits array \(L=\left[l_1,...,l_n \right]\), while the high-bits are stored in a bit-vector $H = \left[h_1,...,h_n \right]$, by concatenating the inverted unary encoding, \(\mathcal{U}\), of the differences (gaps) between consecutive high-bits parts: \(H= \left[\mathcal{U}(h_1-0), \mathcal{U}(h_2-h_1), \dots, \mathcal{U}(h_n-h_{n-1})\right]\). We recall that the inverted unary encoding represents a non-negative integer, \(n\), with \(n\) zeros followed by a one; as an example, 5 is represented by $000001$ (see supplementary Figures 21 and 23 for a more detailed illustration of this scheme). 

The rank and select operations on the Elias-Fano representation require two fundamental operations: finding the \(i\)-th 1 or 0 on a bit-vector. To perform them in an average constant time, having preset a quantum $q$, we build an index for the zeros, $O_0 = [o_1, ..., o_k]$, that stores the position of every $q$ zeros, and an index for the ones, $O_1 = \left[o_1, ..., o_k \right]$, that similarly stores the position of every $q$ ones. 

Thanks to the constructed index, when the \(i\)-th value $v$ must be found, the scan can be started from a position, \(o_j,~\text{for}~j = \left \lfloor \frac{i}{q} \right \rfloor\) that is already near to the i-th $v$. Therefore, instead of scanning the whole high-bits array for each search, we only need to scan the high-bits array from position $o_j$ to position $o_{j+1}$. 

It can be shown that such scans take an average constant time $\mathcal{O}(q)$ at a low expense of the memory complexity, since we need \(\mathcal{O}\left(\frac{n}{q} \log_2 n\right)\) bits for storing the two indexes (Supplementary Information Section 7.1). Indeed, in our implementation we chose \(q = 1024\) which provides good performance at the cost of a low memory overhead of \(3.125\%\) over the high-bits and, on average, for every select operation we need to scan 16 words of memory.

\subsubsection{Available Data-structure Trade-offs}\label{methods:time_space_trade_offs}

\gra~offers a choice between two data structures, Compressed Sparse Row (CSR) and Elias-Fano, at compile time. The CSR data structure is the default option due to its speed and efficiency in handling common graph operations, such as exploring a node's neighbourhood. This structure stores the graph as an array of row pointers, column indices, and non-zero values, providing efficient access to the non-zero elements in sparse adjacency matrices.

On the other hand, Elias-Fano's succinct data structure is primarily effective for representing large graphs because, as mentioned earlier, it requires the least amount of memory without additional assumptions. The Elias-Fano structure is recommended in cases where the graph size is so big that memory conservation becomes crucial.

While \gra~provides the option to choose between two data structures, an expert user can add and use any other graph data structure optimized for their specific task.

\subsection{Spectral and matrix factorization embedding methods} \label{sec:spectral_node_embedding}

Spectral and matrix factorization methods start by computing weighted adjacency matrices and may include one or more factorization steps. Secondly, given a target embedding dimensionality \(k\), these models generally use as embeddings the \(k\) eigenvectors or singular vectors corresponding to spectral or singular values of interest.

A description of the spectral and matrix factorization methods implemented in \gra~is reported in Supplementary Information Section 8.1.

\gra~provides efficient parallel methods to compute the initial weighted adjacency matrix of the various implemented methods, which are computed either as dense or sparse matrices depending on how many non-zero values the metrics are expected to generate. The computation of the singular vectors and eigenvectors are currently computed using the state-of-the-art LAPACK library~\cite{anderson1999lapack}, though more scalable methods that compute the vectors using an implicit representation of the weighted matrices are currently under investigation.

\subsection{First and second-order RW-based embedding methods} \label{sec:random_walks_methods}

First- and second-order random-walk embedding models are shallow neural networks, generally composed of two layers and trained on batches of random-walk samples. Given a window size, these models learn some properties of the sliding windows on the RWs, such as the co-occurrence of two nodes in each window using Glove~\cite{pennington2014glove}, the window central node given the other nodes in the window using CBOW~\cite{Mikolov13b}, or vice-versa the nodes in the window from the window central node using Skipgram~\cite{Mikolov13b}. The optimal window size value may vary considerably depending on the graph diameter and overall topology. Once the shallow model has been optimized, the weights in either the first or the second layer can be used as node embeddings.

An overview of the RW-based methods implemented in \gra~is reported in Supplementary Information Section 8.2.

\paragraph{Efficient implementation of SkipGram and CBOW models.}
\gra~ provides both its own implementations and Keras-based implementations for all shallow neural network models (e.g. CBOW, SkipGram, TransE). Nevertheless, since shallow models allow for particularly efficient data-race aware and synchronization-free implementations~\cite{zhang2017}, the \virgolette{by-scratch} \gra~ implementations significantly outperform the Keras-based ones, as TensorFlow APIs are too coarse and high-level for such fine-grained optimizations. While GPU training is available for the TensorFlow models, their overhead with shallow models tends to be so relevant that \virgolette{by-scratch} CPU implementations outperform those based on GPU. Moreover, the embedding of large graphs (such as Wikipedia) do not fit in most GPU hardware memory. Still, Keras-based models allow users to experiment with the open-software available in the literature for Keras, including, e.g., advanced optimizer and learning rate scheduling methods.

SkipGram and CBOW models are trained using scale-free negative sampling, which is efficiently implemented using the Elias-Fano data structure rank method.

To obtain reliable embeddings, the training phase of the shallow model would need an exhaustive set of random-walk samples to be provided for each source node, so as to fully represent the source-node context. When dealing with big graphs, the computation of a proper amount of random-walk samples needs efficient routines to represent the graph into memory, retrieve and access the neighbors of each node, randomly sample an integer, and, in case of (Node2Vec) second-order RWs~\cite{grover2016node2vec}, compute the transition probabilities, which must be recomputed at each step of the walk.

The first-order RW is implemented using a SIMD routine for sampling integers (Supplementary Information Section 7.2.1). When the graph is weighted, another SIMD routine is used to compute the cumulative sum of the unnormalized probability distribution (Supplementary Information Section 7.2.2). The implementation of the second-order RW requires more sophisticated routines described in sections \ref{methods:efficient_random_walks}, and \ref{sec:our_alias_rethinking}. Moreover, in section \ref{sec:methods:approximated_random_walks} we present an approximated weighted and second-order RW that allows to deal with high-degree nodes.

\subsubsection{Implementation of second-order RWs} \label{methods:efficient_random_walks}
Node2Vec is a second-order random-walk sampling method~\cite{grover2016node2vec}, whose peculiarity relies in the fact that the probability of stepping from one node \(v\) to its neighbours considers the preceding step of the walk (Supplementary Figure 27). More precisely, Node2Vec defines the un-normalized transition probability \(\pi_{vx}\) of moving from \(v\) to any direct neighbor \(x\), starting at a previous step from node $t$, as a function of the weight \(w_{vx}\) on the edge connecting \(v\) and \(x\) \((v, x)\), and a search bias \(\alpha_{pq}(t, x)\):
\[\pi_{vx} = \alpha_{pq}~(t, x)~w_{vx}\] 

The search bias \(\alpha_{pq}(t, x)\) is defined as a function of the distance \(d(t, x)\) between \(t\) and \(x\), and two parameters \(p\) and \(q\), called, respectively, the \textit{return} and \textit{in-out} parameters:
\begin{equation}
	\label{eq:alphaRW}
	\alpha_{pq}(t, x) = \left\{\begin{matrix}
	\frac{1}{p} & {\rm if }\; d(t, x) = 0\\ 
	1 & {\rm if }\; d(t, x) = 1\\ 
	\frac{1}{q} & {\rm if }\; d(t, x) = 2
	\end{matrix}\right.
\end{equation} 

If the return parameter \(p\) is small, the walk will be enforced to return to the preceding node; if \(p\) is large, the walk will otherwise be encouraged to visit new nodes. The in-out parameter \(q\) allows to vary smoothly between Breadth First Search (BFS) and Depth First Search (DFS) behaviour. Indeed, when \(q\) is small the walk will prefer outward nodes, thus mimicking DFS; it will otherwise prefer inward nodes emulating in this case BFS. Since \(\alpha\) must be recomputed at each step of the walk, the algorithm to compute it must be carefully designed to guarantee scalability. 

In \gra~we sped up its computation by decomposing the search bias $\alpha_{pq}(t,x)$ into the in-out bias \(\beta_q(t, x)\), related to the $q$ parameter, and the return bias \(\gamma_p(t, x)\), related to $p$:
\begin{equation}
	\alpha_{pq}(t, x) = \beta_q(t, x) \gamma_p(t, x)
	\label{eq:alpha}
\end{equation}
where the two new biases are defined as:
\begin{equation}
	\label{eq:beta}
	\beta_q(t, x) = \left\{\begin{matrix}
	1 & {\rm if }\; d(t, x) \le 1\\ 
	\frac{1}{q} & {\rm if }\; d(t, x) = 2
	\end{matrix}\right.
	\qquad\qquad
	\gamma_p(t, x) = \left\{\begin{matrix}
	\frac{1}{p} & {\rm if }\; d(t, x) = 0\\
	1 & {\rm if }\; d(t, x) > 0
	\end{matrix}\right.
\end{equation}
It is easy to see that eq.~\ref{eq:alpha} is equivalent to eq.~\ref{eq:alphaRW}.

\paragraph{Efficient computation of the in-out and return bias.}
\label{sec:efficient_search_bias}
The in-out bias can be re-formulated to allow an efficient implementation: starting from an edge \((t, v)\) we need to compute \(\beta_q(t, x)\) for each \(x \in N(v)\), where \(N(v)\) is the set of nodes adjacent to \(v\) including node \(v\) itself.
\[
	\beta_q(t, x) = \left\{\begin{matrix}
	1 & {\rm if }\; d(t, x) \le 1\\ 
	\frac{1}{q} & {\rm otherwise}
	\end{matrix}\right.
	\qquad \Rightarrow \qquad
	\beta_q(t, x) = \left\{\begin{matrix}
	1 & {\rm if }\; x \in N(t)\\ 
	\frac{1}{q} & {\rm otherwise}
	\end{matrix}\right.
\]

This formulation (Supplementary Figure 26) allows us to compute in batch the set of nodes \(X_\beta\) affected by the in-out parameter $q$:
\[
	X_\beta = \left \{ x \mid \beta_q(t, x) = \frac{1}{q} , q \neq 1 \right \} = N(v) \setminus N(t)
\]
where \(N(v)\) are the direct neighbors of node \(v\). In this way, the selection of the nodes $X_\beta$ affected by $\beta_q$ simply require to compute the difference of the two sets \(N(v) \setminus N(t)\).
We efficiently compute $X_\beta$ by using a SIMD algorithm implemented in assembly, leveraging AVX2 instructions that work on node-set representations as sorted vectors of the indices of the nodes (see Supplementary Information Sections 7.2.1 and 7.2.2 for more details). The return bias $\gamma_p$ can be simplified as: 
\[
	\gamma_p(t, x) = \left\{\begin{matrix}
	\frac{1}{p} & {\rm if }\; d(t, x) = 0\\
	1 & {\rm otherwise}
	\end{matrix}\right.
	\qquad\Rightarrow\qquad
	\gamma_p(t, x) = \left\{\begin{matrix}
	\frac{1}{p} & {\rm if }\; t = x\\
	1 & {\rm otherwise}
	\end{matrix}\right.
\]
It can be efficiently computed using a binary search for the node \(t\) in the sorted vector of neighbours. Summarizing, we re-formulated the transition probability $\pi_{vx}$ of a second-order RW in the following way:
\[
	\pi_{vx} = \beta_q(t, x) \gamma_p(t, v, x)w_{vx}
	\qquad
	\beta_q(t, x) = \left\{\begin{matrix}
	1 & {\rm if }\; x \in N(t)\\ 
	\frac{1}{q} & {\rm otherwise}
	\end{matrix}\right.
	\qquad
	\gamma_p(t, v, x) = \left\{\begin{matrix}
	\frac{1}{p} & {\rm if }\; t = x\\
	1 & {\rm otherwise}
	\end{matrix}\right.
\]

If \(p\), \(q\) are equal to one, the biases can be simplified, so that we can avoid computing them. In general, depending on the values of $p, q$ and on the type of the graph (weighted or unweighted), \gra~provides eight specialized implementation of the Node2Vec algorithm, to significantly speed-up the computation (Supplementary Tables 50 and 51). \gra~automatically selects and runs the specialized algorithm that corresponds to the choice of the parameters $p, q$ and the graph type. This strategy allows a significant speed-up. For instance, in the base case ($p=q=1$ and an unweighted graph) the specialized algorithm runs more than 100 times faster than the most complex one ($p \neq 1, q \neq 1$, weighted graph). Moreover, as expected, we observe that the major bottleneck is the computation of the in-out bias (Supplementary Table 51).

\subsubsection{Efficient sampling for Node2Vec RWs}
\label{sec:our_alias_rethinking}

Sampling from a discrete probability distribution is a fundamental step for computing a RW and can be a significant bottleneck. Many graph libraries implementing the Node2Vec algorithm speed up sampling by using the Alias method (see Supplementary Information Section 7.2.3), which allows sampling in constant time from a discrete probability distribution with support of cardinality \(n\), with a pre-processing phase that scales linearly with \(n\). 

The use of the Alias Method for Node2Vec incurs the \virgolette{memory explosion problem} since the preprocessing phase for a second-order RW on a graph with \(\abs{E}\) edges has a support whose cardinality is \(\mathcal{O}\left(\sum_{e_{ij} \in E} \text{deg}\left(j\right)\right)\), where \(\text{deg}(j)\) is the degree of the destination node of the edge \(e_{ij} \in E\). 

Therefore, the time and memory complexities needed for preprocessing make the Alias method impractical even on relatively small graphs. For instance, on the unfiltered Human STRING PPI graph (\(19.354\) nodes and \(5.879.727\) edges) it would require \(777\) GB of RAM. 

To avoid this problem, we compute the distributions on the fly. 
For a given source node \(v\), our sampling algorithm applies the following steps: 
\begin{enumerate}
\item computation of the un-normalized transition probabilities to each neighbour of \(v\) according to the provided \textit{in-out} and \textit{return} biases;
\item computation of the un-normalized cumulative distribution, which is equivalent to a cumulative sum;
\item uniform sampling of a random value between 0 and the maximum value in the un-normalized cumulative distribution;
\item identification of the corresponding index through either a linear scan or a binary search, according to the degree of the node \(v\).
\end{enumerate}

To compute the cumulative sum efficiently, we implemented a SIMD routine that processes at once in CPU batches of 24 values. Moreover, when the length of the vector is smaller than 128, we apply a linear scan instead of a binary search because it is faster thanks to lower branching and better cache locality. Further details are available in the Supplementary Information Section 7.2.2.

\subsubsection{Approximated RWs} \label{sec:methods:approximated_random_walks}
Since the computational time complexity of the sampling algorithm for either weighted or second-order RWs scales linearly with the degree of the considered source node, computing an exact RW on graphs with high degree nodes (where "high" refers to nodes having an outbound degree larger than \(10000\)) would be impractical, also considering that such nodes have a higher probability to be visited. 

To cope with this problem, we designed an approximated RW algorithm, where each step of the walk considers only a sub-sampled set of \(k\) neighbors, where the parameter \(k\) is set to a value significantly lower than the maximum node degree. 

An efficient neighborhood sub-sampling for nodes with degree greater than \(k\) requires to uniformly sample unique neighbors whose original order must be maintained. To uniformly sample distinct neighbors in a discrete range \([0, n]\) we developed an algorithm (Sorted Unique Sub-Sampling - SUSS)  that divides the range \([0, n]\) into \(k\) uniformly spaced buckets and then randomly samples a value in each bucket. 
The implementation of the algorithm is reported in Supplementary Algorithm 1 (Supplementary Information Section 7.2.4). After splitting the range $[0, \dots, n-1]$ into $k$ equal segments (buckets) with length \(\left\lfloor\text{delta} / k \right\rfloor \), SUSS samples an integer from each bucket by using Xorshift random number generator.  To establish whether the distribution of the integers sampled with SUSS is truly approximating an uniform distribution, we  sampled $n=10.000.000$ integers over $[0,\ldots,10.000]$, by using both SUSS and by drawing from a uniform distribution in  $[0,\ldots,10.000]$. We then used the one-sided Wilcoxon signed-rank test to compare the frequencies of the obtained indices and we obtained a p-value of \(0.9428\), meaning that there is not a statistically significant difference among the two distributions. Therefore, by using a time complexity \(\Theta(k)\) and a spatial complexity \(\Theta(k)\) SUSS produces reliable approximations of a uniform distribution.


The disadvantage of this sub-sampling approach is that two consecutive neighbors will never be selected in the same sub-sampled neighborhood. 
Nevertheless considering that the sub-sampling is repeated at each step of the walk, consecutive neighbors have the same probability of being selected in different sub-samplings.


\subsection{Triple-sampling and corrupted triple sampling methods} \label{sec:triples_sampling_corrupted_sampling}
Triple sampling methods are shallow neural networks trained on triples, \((v, \ell, s)\), where \(\left\{v, s\right\}\) is a node-pair composed of a source (\(v\)) and a destination node (\(s\)), and \(\ell\) is a property of the edge \((v, s)\) connecting them.
Similar to triple sampling methods, \textit{corrupted}-triple sampling methods are trained on the (\textit{true}) triples \((v, \ell, s)\), but also on \textit{corrupted triples}, that are obtained by corrupting the original triples by substituting the source and/or destination nodes \(\left\{v, s\right\}\) with randomly sampled nodes \(\left\{v', s'\right\}\), while maintaining the attribute unchanged \((v', \ell, s')\).
More details about triple-sampling and corrupted triple-sampling methods are available in Supplementary Information Section 8.3 and 8.4.

\gra~provides a full implementation of first and second-order LINE triple sampling methods~\cite{tang2015line}, as well as a Rust parallel implementation of the TransE corrupted triple sampling method~\cite{zhang2017}.
Moreover, a large set of corrupted-triple sampling models is integrated from the PyKeen library. The integrated models include TransH, DistMult, HolE, AutoSF, TransF, TorusE, DistMA, ProjE, ConvE, RESCAL, QuatE, TransD, ERMLP, CrossE, TuckER, TransR, PairRE, RotatE, ComplEx, and BoxE~\cite{ali2021pykeen}. We refer to each of the original papers for the extensive explanation. The parameters used for the evaluation of node embedding models in \gra~pipelines are available in the Supplementary Information Section 4.1.

\subsection{Edge embedding methods and graph visualization} \label{methods:edge_embedding_models}

\gra~offers an extensive set of methods to compute edge embeddings from node embeddings (e.g. concatenation, average, cosine distance, L1, L2 and Hadamard operators \cite{grover2016node2vec}), and the choice of the specific edge-embedding operator is left to the user, who can set it through a parameter. To meet the various model requirements, the library provides three implementations of the edge embedding. In the first one, all edge embedding methods are implemented as Keras/TensorFlow layers and may be employed in any Keras model. In the second one, all methods are also provided in a NumPy implementation. Finally, a third one uses Rust for models where performance is particularly relevant. For instance, the cosine similarity computation in the Rust implementation is over \(250\times\) faster than the analogous NumPy implementation. Whenever possible, the computation of edge embeddings is executed lazily for a given subset of the edges at a time since the amount of RAM required to explicitly rasterize the edge embedding can be prohibitive on most systems, depending on the edge set cardinality of the considered graph. More specifically, while the lazy generation of edge embeddings is possible during training for only a subset of the supported edge and edge-label prediction models, it is supported for all models during inference.

The library also comes equipped with tools to visualize the computed node and edge embedding and their properties, including edge weights, node degrees, connected components, node types and edge types. For example, in figure \ref{fig:schema} \textbf{c} we display the node (left) and edge types (center) of the KG-COVID19 graph and whether sampled edges exist (right) by using the first two components of the \tsne~decomposition of the node/edge embeddings~\cite{van2008visualizing}.

\subsection{Node-label, edge-label, and edge prediction models} \label{methods:prediction_models}

\gra~provides implementations to perform node-label prediction, edge-label prediction and edge prediction tasks.

All the models devoted to any of the three prediction tasks share the following implementation similarities. 
Firstly, they all implement the \textit{abstract classifier interface} and therefore provide straightforward methods for training (\textit{fit}) and inference (\textit{predict} and \textit{predict\_proba}). 

Secondly, all models are multi-modal, that is, they not only can receive the (user-defined) node/edge embedded representation, but also other embeddings computed in multiple ways and therefore carrying different semantics (e.g., topological node/edge embeddings or BERT embeddings). For edge prediction and edge-label prediction models, this also generalizes to multiple node-type features, which, if available, are concatenated to the considered node features, and to the possibility of computing traditional edge metrics (e.g. Jaccard, Adamic-Adar, and so on).

For each task, we make available at least eight models from the literature, adapted to the considered task: \(5\) are Scikit-learn-based models, namely Random Forest, Extra Trees, Decision Tree, Multi-Layer Perceptron (MLP), and Gradient Boosting. The remaining \(3\) are TensorFlow-based models, namely GraphSAGE~\cite{hamilton2020graph}, Kipf GCN~\cite{welling2016semi} and a baseline GNN. 

As per the node embedding models, custom and third party models can be integrated through task-specific Python abstract classes (Section \ref{sec:automated_pipeline}).

Scikit-learn-based models make available all the parameters that are available in the Scikit version. 
It is straightforward to achieve a significant speed-up without any modification of the scikit-learn code by simply using Intel's sklearnex~(\url{https://www.intel.com/content/www/us/en/developer/articles/technical/benchmarking-intel-extension-for-scikit-learn.html}).

TensorFlow-based models make available parameters to set the number of layers in each provided feature's sub-module and head module. 

Visualizations of the Kipf GCN model for node-label, edge-label and edge prediction tasks are also available (see Supplementary Information Section 9).

All edge prediction models can be trained by sampling the graph negative edges by either following a uniform or a scale-free distribution; by default we set a scale-free distribution because it generally produces more informative negative-training sets, characterized by a smaller covariate-shift with respect to the positive-set. This approach still guarantees a negligible number of false negatives edges.
The unbalance between positive and negative edges is also a free parameter which may be arbitrarily set: by default the models are trained using a balanced approach, that is we sample a number of negative edges equal to the number of positive edges.

In addition to the eight models presented in section \ref{methods:prediction_models}, we also make available a multi-modal perceptron model implemented in Rust. This model, analogously to all other models, supports lazy computation of edge embedding and edge features, but does this in an extensively parallel manner with no additional memory requirement over the model weights. The model optimizer is Nadam. The Perceptron is a great baseline for comparison, given its rapid convergence, minimal hardware requirements (no GPUs nor significant RAM requirement), and competitive performance in many considered tasks. Such a model is essential to put into perspective the improvements achieved by more complex and often significantly more expensive models.

Parameters used for the evaluation of edge prediction models in \gra~pipelines are available in the Supplementary Information Section 4.2.

All of the provided edge-label prediction models support binary and multi-class classification tasks. We currently lack support for multi-label classification tasks, which is being addressed.

All of the provided node-label prediction models support binary, multi-class and multi-label classification tasks. Parameters used for the evaluation of node-label prediction models in \gra~pipelines are available in the Supplementary Information Section 4.3.

\subsection{Pipelines for the evaluation of graph-prediction tasks} \label{sec:automated_pipeline}
To provide actionable and reliable results, the fair and objective comparative evaluation of datasets, graph embedding, and prediction models is crucial and not only requires specifically designed and real-world benchmark datasets~\cite{hu2020open}, but also pipelines that could allow non-expert users to easily test and compare graphs and inference algorithms on the desired graphs.

\subsubsection{FAIR graph retrieval} 
\label{sub:grape-fair-data}

\gra~facilitates \textit{FAIR} access to an extensive set of graphs and related datasets, including both commonly used benchmark datasets and graphs actively used in biomedical research. Any of the available graphs can be retrieved and loaded with a single line of Python code (Fig.~\ref{fig:schema} \textbf{b.}), and their list is constantly expanding, thanks to the generous contributions of \gra~users. The list of resources currently supported can be found at Supplementary Information Section 3.1.

\paragraph{Findability and Accessibility.} Datasets may change locations, versions may appear in more than one location, and file formats may change. Using an ensemble of custom web scrapers, we collect, curate and normalize the most up-to-date datasets from an extensive resources list (currently over \(80,000\) graphs). The collected metadata is shipped with each \gra~release, ensuring end-users can always find and immediately access any available version of the provided datasets.

\paragraph{Interoperability.} The graph retrieval phase contains steps that robustly convert data from (even malformed) datasets into general-use TSV documents that, while primarily used as graph data, can be used for any desired application case.

\paragraph{Reusability.} Once loaded, the graphs can be arbitrarily processed and combined, used with any of the many embedding and classifier models from either the \gra~library or any third-party model integrated in \gra~by implementing the interface described in section \ref{sub:grape-fair-comparison-pipeline}.

\subsubsection{FAIR evaluation pipelines} 
\label{sub:grape-fair-comparison-pipeline}

\gra~provides pipelines for evaluating node-label, edge-label and edge prediction experiments trained on user-defined embedding features and by using task-specific evaluation schemas.



In particular, the evaluation schemas for edge prediction models are K-fold cross-validations, Monte Carlo, and Connected Monte Carlo (Monte Carlo designed to avoid the introduction of new connected components in the training graph) holdouts. All of the edge prediction evaluation schemas may sample the edges in a uniform or stratified way, with respect to a provided list of edge-types. Sampling of negative (non-existing) edges may be executed by either following a uniform or a scale-free distribution. Furthermore, the edge-prediction evaluation may be performed by using varying unbalance ratios (between existent and non-existent edges) to better gauge the true-negative rate (specificity) and false-positive rate (fall-out). Stratified Kfold and stratified Monte Carlo holdouts are also provided for node and edge-label prediction models. 

For all tasks, an exhaustive set of evaluation metrics are computed, including AUROC, AUPRC, Balanced Accuracy, Miss-rate, Diagnostic odds ratio, Markedness, Matthews correlation coefficient and many others.


All the implemented pipelines have integrated support for differential caching, storing the results of every step of the specific experiment, and for \virgolette{smoke tests}, i.e. for running a lightweight version of the experimental setup with minimal requirements to ensure execution until completion before running the full experiment. 

The pipelines can use any model implementing a standard interface we developed. The interface requires the model to implement methods for training (\textit{fit} or \textit{fit\_transform}), inference (\textit{predict} and \textit{predict\_proba}) plus additional metadata methods (e.g., whether to use node types, edge types, and others) which are used to identify experimental flaws and biases. As an example, in an edge-label prediction task using node embeddings, \gra~will use the provided metadata to check whether the selected node embedding method also uses edge labels. If so, the node embedding will be recomputed during each holdout. Conversely, if the edge labels are not used in the node embedding method, it may be computed only once. The choice to recompute the node embedding for each holdout, which may be helpful to gauge how much different random seeds change the performance, is left to the user in this latter case.

To configure one of the comparative pipelines, users have to import the desired pipeline from the \gra~library and specify the following modular elements:

\begin{description}
    \item[Graphs] The graphs to evaluate, which can be either graph objects or strings matching the names from graphs retrieval.
    \item[Graph normalization callback] For some graphs it is necessary to execute normalization steps and filtering, such as the STRING graphs which can e.g. to be filtered at \(700\) minimum edge weight. For this reason, users can provide this optional normalization callback.
    \item[Classifier models] The classifier models to evaluate, which can either be a model implemented in \gra~or custom models implementing the proper interface.
    \item[Node, node type, and edge features] The features to be used to train the provided classifier models. These features can be node embedding models, either implemented in \gra~or custom embedding models implementing the node embedding interface.
    \item[Evaluation schema] The evaluation schema to follow for the evaluation.
\end{description}

Given any input graph, each pipeline starts by retrieving it (if the name of the graph was provided) and validating the provided features (checking for NaNs, constant columns, compatibility with the provided graphs); next, and if requested by the user, it computes all the node-embeddings to be used as additional features for the prediction task. 
Once this preliminary phase is completed, the pipeline starts to iterate and generate holdouts following the provided evaluation schema. 

For each holdout, \gra~then computes the node embeddings required to perform the prediction task (such as topological node embeddings for a node-label prediction task, or topological node embeddings followed by their combination through a user-defined edge embedding operator - see Section \ref{methods:edge_embedding_models} - to obtain the edge embedding in an edge-prediction task), so that a new instance of the provided classifier models can be fitted and evaluated (by using both the required embedding and, eventually, the additional, label-independent, features computed in the preliminary phase). The classifier evaluation is finally performed by computing an exhaustive set of metrics including AUROC, AUPRC, Balanced Accuracy, Miss-rate, Diagnostic odds ratio, Markedness, Matthews correlation coefficient and many others.

Interfaces are made available for \href{https://github.com/monarch-initiative/embiggen/blob/develop/embiggen/utils/abstract_models/abstract_embedding_model.py}{embedding models}, \href{https://github.com/monarch-initiative/embiggen/blob/develop/embiggen/node_label_prediction/node_label_prediction_model.py}{node-label prediction}, \href{https://github.com/monarch-initiative/embiggen/blob/develop/embiggen/edge_label_prediction/edge_label_prediction_model.py}{edge-label prediction}, and \href{https://github.com/monarch-initiative/embiggen/blob/develop/embiggen/edge_prediction/edge_prediction_model.py}{edge prediction}. All models available in \gra~implement these interfaces, and they can be used as starting points for custom integrations. Many usage examples  are available in the library tutorials: \href{https://github.com/AnacletoLAB/grape/tree/main/tutorials}{ https://github.com/AnacletoLAB/grape/tree/main/tutorials}.

\section*{Data availability}
\gra~graph retrieval includes all the graphs used in the \ens~benchmarks and the pipeline experiments and are all available from \url{https://github.com/AnacletoLAB/grape}. 
Graphs used for the Ensmallen benchmarks are detailed in Supplementary Information Section 2. Graphs used for edge and node-label prediction experiments are detailed in Supplementary Information Section 3.
The real world graphs used in Section~\ref{sub:grape-scaling}
are downloadable from  \url{https://archive.org/download/ctd_20220404/CTD.tar} (Pre-built CTD),  \url{https://archive.org/download/pheknowlator_20220411/PheKnowLator.tar} (Pre-built biomedical PheKnowLator data), and  \url{https://archive.org/download/wikipedia\_edge\_list.npy/wikipedia\_edge\_list.npy.gz} (Pre-built English Wikipedia).
More details are available in Supplementary Information  Section 6. 
The procedures for the construction of train and test graphs for edge prediction are detailed in Supplementary Information  Section 10.2.
Source Data for Figures 2, 3, 4, and 5 are available through this manuscript.

\section*{Code availability}
All the code of the experiments presented within the manuscript are publicly available from GitHub repositories. \gra~can be installed from PyPI: \url{https://pypi.org/project/grape}. 
The source code, reference manual and tutorials for its usage, alongside several application examples, are available on GitHub: \url{https://github.com/AnacletoLAB/grape}. In particular more than $50$ tutorials to learn how to use the main functionalities of~\gra~are available from \url{https://github.com/AnacletoLAB/grape/tree/main/tutorials}.

All the scripts to reproduce the experiments showed in the paper are available from GitHub:
\begin{itemize}
    \item \ens~benchmarks: loading graphs, executing first and second-order RWs: \url{https://github.com/LucaCappelletti94/ensmallen_experiments}
    \item Approximated RW experiments: \url{https://github.com/AnacletoLAB/grape/blob/main/tutorials/Comparing\%20DeepWalk\%20and\%20Node2Vec\%20running\%20on\%20exact\%20and\%20approximated\%20random\%20walks.ipynb}
    \item Experimental comparison of node embedding methods
        \subitem (a) Edge prediction experiments: \url{https://github.com/AnacletoLAB/grape/blob/main/tutorials/Using\%20the\%20edge\%20prediction\%20pipeline.ipynb}
        \subitem (b) Node-label prediction experiments: \url{https://github.com/AnacletoLAB/grape/blob/main/tutorials/Using\%20the\%20node-label\%20prediction\%20pipeline.ipynb}
    \item Comparison of GRAPE with state-of-the-art libraries on big real-world graphs: \url{https://github.com/LucaCappelletti94/embiggen_experiments/tree/master/node2vec\_comparisons}
\end{itemize}

The software is delivered under the MIT license.



\section*{Acknowledgements}
This research was supported by the "National Center for Gene Therapy and Drugs based on RNA Technology", PNRR-NextGenerationEU program [G43C22001320007], NIH/National Cancer Institute [U01-CA239108-02],
Transition Grant Line 1A Project "UNIMI PARTENARIATI H2020" [PSR2015-1720GVALE\_01], the Common Fund, Office of the Director, National Institutes of Health [U01-CA239108-02], the Monarch Initiative, National Institute of Health [1R24OD011883-01], Project PID2021-128970OA-I00 by MCIN/AEI/10.13039/501100011033/ FEDER and
from the Director, Office of Science, Office
of Basic Energy Sciences of the U.S. Department of Energy under U.S. Department of Energy Contract No. DE-AC02-05CH11231.
The funders had no role in study design, data collection and analysis, decision to publish or preparation of the manuscript.

\section*{Authors' Contributions}
Conceptualization and Methodology: L.C., T.F., G.V., E.C., J.R. and P.N.R.
Software - Design and implementation: L.C., T.F. with the contribution of V.R., T.C. and J.R.
Software - documentation: L.C., T.F.
Data Curation and Investigation: J.R., P.N.R., V.R. and T.C.
Supervision: G.V., E.C., P.N.R., J.R.
Funding Acquisition: C.M., P.N.R., G.V. 
Writing - Original Draft Preparation: G.V., E.C., J.R., P.N.R. 
Writing - Review \& Editing: all authors.

\section*{Competing interests}
The authors declare that they have no competing interests.





\normalem

\end{document}